%% file: main.tex
\pdfoutput=1

\documentclass[11pt]{article}

\usepackage{acl}

\usepackage{times}
\usepackage{latexsym}

\usepackage[T1]{fontenc}

\usepackage[utf8]{inputenc}

\usepackage{microtype}

\usepackage{hyperref}       
\usepackage{url}            
\usepackage{booktabs}       
\usepackage{amsfonts}       
\usepackage{nicefrac}       
\usepackage{microtype}      
\usepackage{xcolor}         
\usepackage{algorithm}
\usepackage{algorithmic}
\usepackage[fleqn]{amsmath}
\usepackage{natbib} 
\usepackage{caption}
\usepackage{subcaption}
\usepackage{gensymb}
\usepackage{amssymb,latexsym}
\usepackage{wrapfig}
\usepackage{todonotes}
\usepackage{tabularx} 
\usepackage{multirow}

\usepackage{mathtools, nccmath}

\usepackage{makecell} 

\input{math_commands}

\newcommand{\leqnomode}{\tagsleft@true\let\veqno\@@leqno}%
\newcommand{\reqnomode}{\tagsleft@false\let\veqno\@@eqno}%
\newcommand*{\compress}{\@minipagetrue}

\newcommand{\slide}{\textsc{SW}}
\newcommand{\bigbird}{\textsc{SW+Rand}}
\newcommand{\distance}{\textsc{Distance}}
\newcommand{\hybrid}{\textsc{Distance+SW}}

\definecolor{yy}{RGB}{220,220,0}
\definecolor{gg}{RGB}{45,190,45}

\makeatletter
\def\BState{\State\hskip-\ALG@thistlm}
\makeatother

\title{Understanding Long Documents with Different Position-Aware Attentions}

\author{
Hai Pham$^{\dagger}$\thanks{\hspace{1mm} Work done during an internship at Microsoft.},\,\,  Guoxin Wang$^{\ddagger}$, Yijuan Lu$^{\ddagger}$, Dinei Florencio$^{\ddagger}$, Cha Zhang$^{\ddagger}$\\
  $^{\dagger}$\textit{Language Technologies Institute, Carnegie Mellon University} 
\\
$^{\ddagger}$\textit{Microsoft Azure AI} \\
  \texttt{htpham@cs.cmu.edu} \ \{\texttt{guow,yijlu,dinei,chazhang}\}\texttt{@microsoft.com}\\
}

\begin{document}
\maketitle

\begin{abstract}
Despite several successes in document understanding, the practical task for long document understanding is largely under-explored due to several challenges in computation and how to efficiently absorb long multimodal input. Most current transformer-based approaches only deal with short documents and employ solely textual information for attention due to its prohibitive computation and memory limit. To address those issues in long document understanding, we explore different approaches in handling 1D and new 2D position-aware attention with essentially shortened context. Experimental results show that our proposed models have the advantages for this task based on various evaluation metrics. Furthermore, our model makes changes only to the attention and thus can be easily adapted to any transformer-based architecture. 
\end{abstract}

\section{Introduction}


The task of document understanding has recently gleaned many successes~\cite{xu2020layoutlm,xu2021layoutxlm,appalaraju2021docformer}. This task requires multimodal input that makes it heavier than the text-only ones, resulting in most models only being capable of dealing with short documents, i.e. having up to 512 tokens. However, there exist long documents almost everywhere, e.g. contracts, scientific papers, newsletters, or Wikipedia articles, which are typically longer than 1,000 words. To automatically summarize and understand such long documents urges long document understanding to become an important task in both natural language processing and artificial intelligence.

\begin{figure}[t]
    \centering
\label{fig:len_distribution}
  \includegraphics[width=0.45\textwidth]{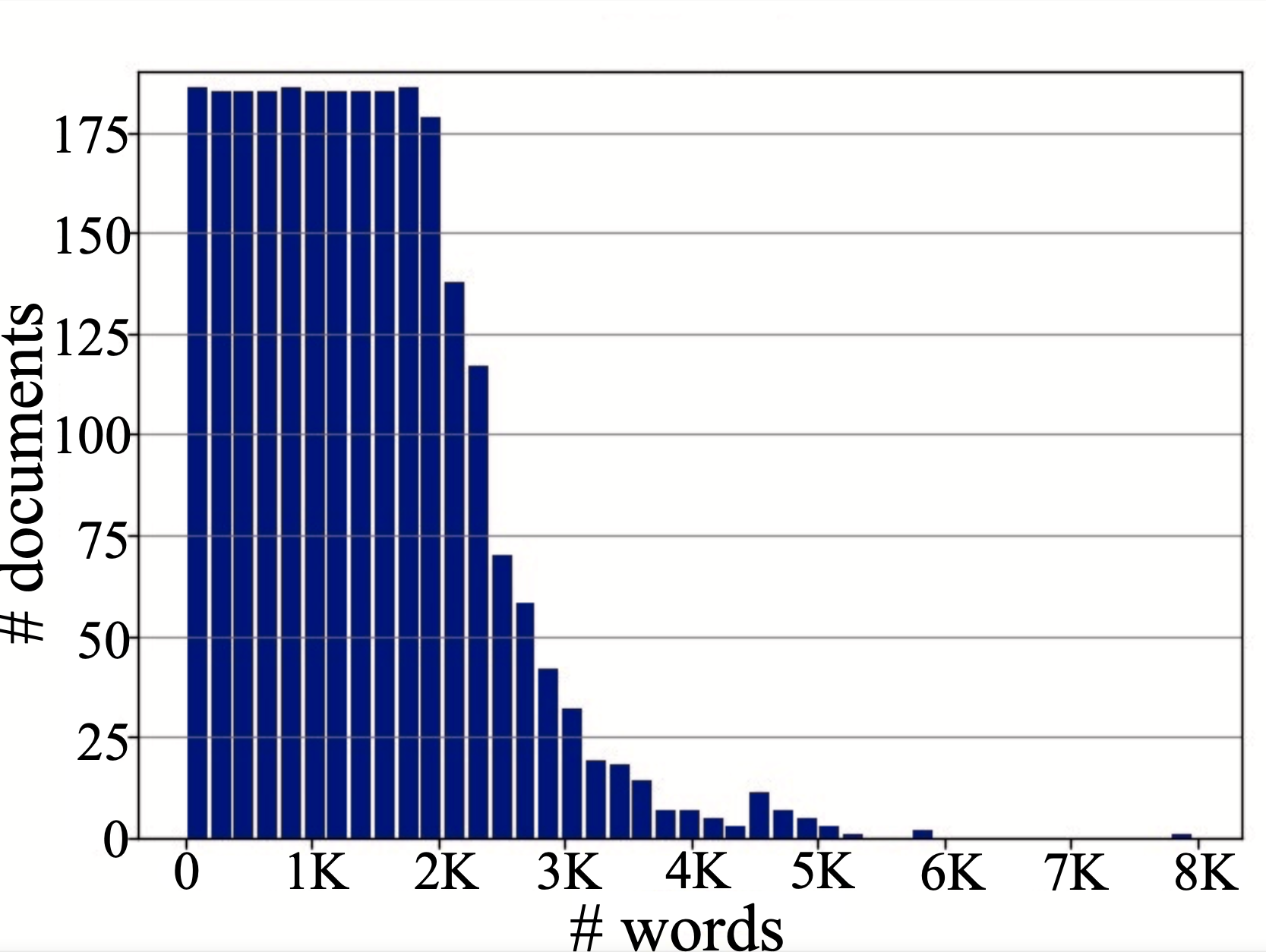}
  \caption{Distribution of document length in RVL-CDIP~\cite{harley2015icdar}, a subset of IIT-CDIP used predominantly in the document understanding pretraining tasks.
  Most of them are longer than 512 words, the limit most current document understanding models accept. We argue that helpful content should span across entire documents and that only accepting 512 words would degrade the performance. 
  }
\end{figure} \label{fig:ldistribution} 

Long document understanding faces several big challenges. 1) Recent document understanding approaches heavily rely on transformer~\cite{vaswani2017attention}. 
However, transformer suffers from the quadratic attention that usually limits the input to 512 words. 
Therefore, the correlation across long paragraphs/pages is yet to be learned. 
2) Understanding long documents requires the power to model all long information available, not only just in text but also in other modalities such as spatial information. 
For example, LayoutLM~\cite{xu2020layoutlm} showed that short document understanding is largely improved by additionally embedding spatial into text information. How to efficiently make use of spatial information for long document understanding, however, is still an open and challenging problem regarding computation cost and adaptability.

Given the fact that long documents frequently appear in practice as well as in many datasets as shown in Figure~\ref{fig:ldistribution}, it is reasonable to assume that useful information is spanned across their lengths. 
Especially current OCR technology, which is essential for data preprocessing, only supports extracting spatial information on every page basis, without the knowledge of other pages. This behavior poses yet another big challenge in dealing with long documents, which requires a proper method to connect information across pages for all input modalities. 

In this paper, we discover new approaches to dealing with long document understanding, which addresses the aforementioned challenges. 
We carefully preprocess OCR data to establish the proper linkages across pages. Then we explore approaches for directly reducing the heavy attention cost while achieving high performance, flexibly using the typical 1D (textual) and/or novelly, 2D (spatial) reduced contextual information, without the need of adding more components into the already-heavy transformer~\cite{appalaraju2021docformer,nguyen2021skim}, employing additional pretraining tasks for better representation learning~\cite{huang2022layoutlmv3,li2021structurallm} or employing complicated new encoding techniques~\cite{hong2021bros,wang2022lilt}.
Despite being simple, we show through experiments that both 1D and 2D information can enhance the practicality of transformer-based models while achieving the needed power of handling long documents without introducing any new pretraining tasks other than the popular one: masked language modeling. 

\textbf{Our contributions} \ In summary, we have three following contributions. 
1) We newly motivate the simplistic, flexible use of spatial input into self-attention, making it plug-able to transformer-based and other architectures using attention. 
2) We are able to tackle the document understanding task with input data up to 4096 words with several attention configurations. 
3) Experimental results prove the advantages of our approaches on various long-document datasets in comparison to short models for both 1D and 2D contextual information. 


\begin{figure*}[h!]
    \centering
\label{fig:len_distribution}
  \includegraphics[width=0.9\textwidth]{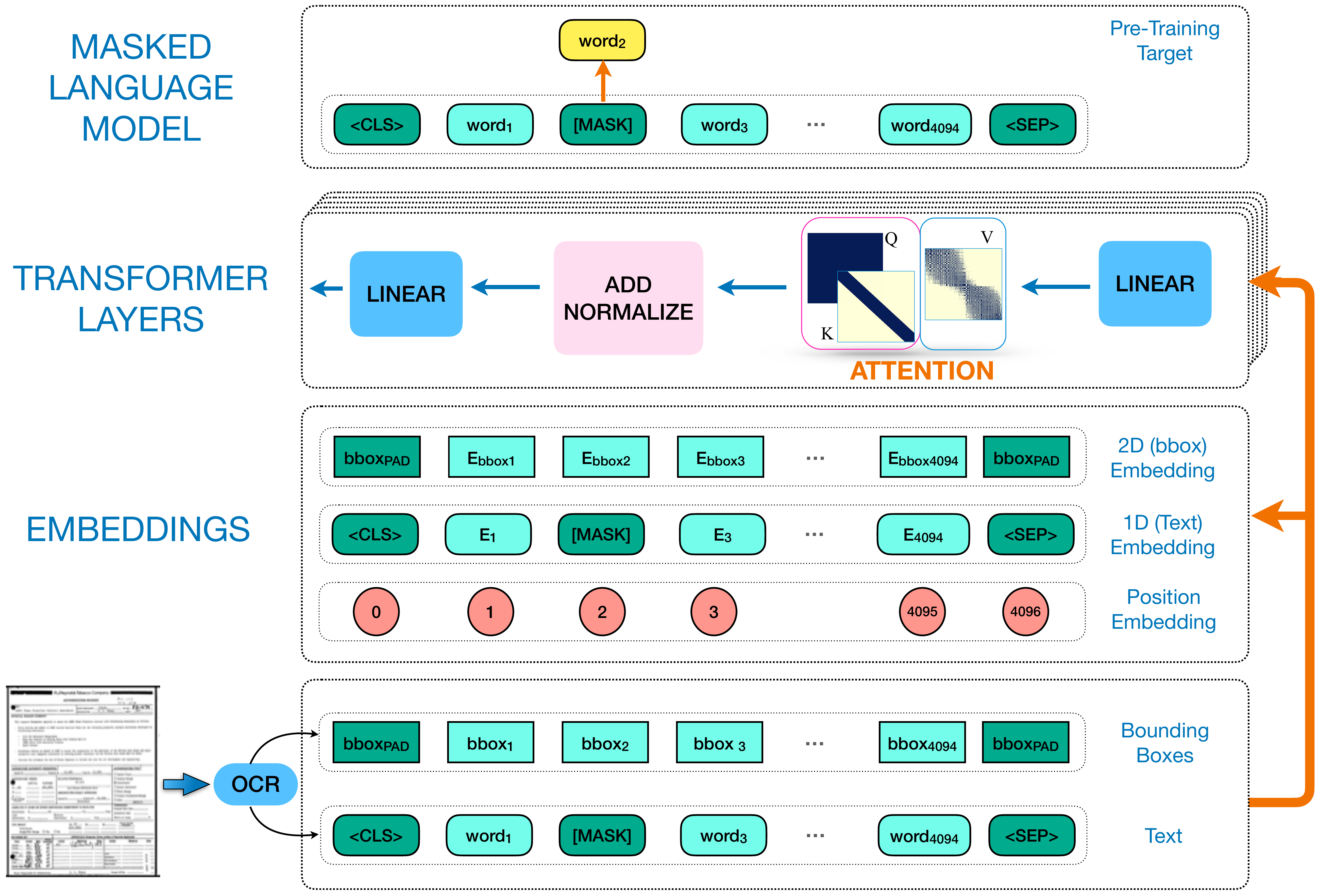}
  \caption{Our pretrain model architecture. Unlike other models for this task, we keep a simple approach by only employing a single MLM pretraining objective and do not employ extra overhead into multimodal embedding or encoding methods. Instead, we tackle the attention module directly and make necessary changes to deal with our focus on long documents, by flexibly using 1D and 2D input. 
  }
\end{figure*} \label{fig:architecture}

\section{Related Work} \label{sec:related} 

\textbf{Transformer Attention For Long Documents}
There are several methods that address the quadratic cost of the transformer attention and some of them narrow the focus on long documents for their practicality. 
Longformer~\cite{beltagy2020longformer} uses sliding windows to reduce the context, only retrains some sparse global connections. 
Similarly, ETC~\cite{ainslie2020etc} embeds relative positions and adds contrastive predictive encoding. 
Bigbird~\cite{zaheer2020big} adds a few random connections on top of the sliding windows and sparse global connections and then arranges a long context into a few blocks to reduce the number of intermediate matrix re-arrangement and calculation steps.
Our model similarly uses sliding windows to effectively handle long documents but differs in that it addresses the complication of multimodal, instead of text-only, data and exploits layout input along with the typical text input flexibly and directly into attention and thus enhancing the attention more power and flexibility in dealing with different data types.

\textbf{Multimodal Document Pretraining} 
Document understanding largely inherits from multimodal pretraining~\cite{li2020oscar,chen2020uniter,luo2020univilm} with the successes from LayoutLM~\cite{xu2020layoutlm,xu2020layoutlmv2}. 
Docformer~\cite{appalaraju2021docformer} and StrucuturalLM~\cite{li2021structurallm} developed the task further by introducing a new two-pronged approach: having new pretraining tasks and suitable changes to the processing or embedding. 
Similarly, LayoutLMv3~\cite{huang2022layoutlmv3} introduces two new, additional pretraining tasks on top of masked language modeling (MLM) to enrich the representation learned by the models. 
Yet another approach is to focus on encoding the spatial information properly with either relative spatial encoding~\cite{hong2021bros} or having separate encoding flows for textual and spatial input, then flexibly fusing them~\cite{wang2022lilt}. Unlike all of those approaches, our solution has a different focused motivation that is long documents, only employs MLM as the only pretraining objective, and tackles the attention directly--by efficiently handling the shortened contexts based on textual/spatial information to deal with long contexts--instead of resorting to further embedding and/or encoding all information properly, resulting in a more simple and lightweight solution that can be adapted easily for any architecture using the attention mechanism.

Finally, Skim-Attention~\cite{nguyen2021skim} probably has the most related motivation for long documents, although we have a more memory-efficient, and faster way of handling layout input directly into attention and not from after the embedding like theirs, and consequently support longer input (4096 vs. 2048).  


\section{Our Model} \label{sec:model}

The structure of this section is as follows. We will first introduce our MLM pretraining model with an emphasis on the novel attention that employs the direct flexible use of either textual (1D) or spatial (2D) information. Next, we explain the post-processing of OCR and its crucial importance in MLM for document intelligence. Then we explain different attention configurations based on 1D and 2D inputs. Finally, we enumerate the models associated with those attention modules. 



\subsection{Pretrain Model Architecture}

To keep our solution simplistic and easy for studying the effects of each approach being proposed, we only employ Masked Language Model (MLM) architecture as in other document intelligence work, e.g. ~\citet{xu2020layoutlm,xu2020layoutlmv2}. However, we discover new attention approaches in MLM to enable its capability of handling long documents. 
In more detail, different from a typical MLM predominantly used in natural language processing, we have multimodal--instead of text-only--input, which inevitably makes the model heavier and hence cannot deal with long documents without proper changes, as we propose below.  



First, we use the sliding-window inspired from~\citet{beltagy2020longformer}, given its lightweight and elegance in limiting the context window, making it significantly more memory friendly. 
Second, we introduce new spatial-based attention masks, in which each context window to a bounding box is determined by calculating its spatial neighbors, instead of the given neighboring words. 
Likewise, our model not only uses spatial input in the embedding but also in attention directly with preserved spatial correlation. 
The illustration of our MLM model is shown in Figure~\ref{fig:architecture}. Additionally, section~\ref{subsec:distance_mask} will elaborate on the establishment and usage of these new distance masks in comparison with others. 

\subsection{Post-OCR Processing}

The task of document intelligence relies heavily on the quality of the OCR pre-processing as the first data processing. As a result, how to present the post-OCR data properly to the model is very important, as any mistake in this phase will be compounded later in the model. 
Especially in the case of long documents, this processing is more crucially important. While long documents have multiple pages, current OCR engines only generate single-page results, without any connections among pages. 
More current models are ``short'' models that support up to 512 tokens, and thus typically make use of the very first page's OCR results, discarding the rest of the valuable information. As a result, the further need for post-processing is usually unnecessary in those models.

Unlike those short models, to make our model capable of tackling long documents, we process and normalize the post-OCR data 
to establish the connections for all input components among the pages. 
For example, the bounding boxes on page $n$ need adjusting the coordinates to include the previous $n-1$ pages. 

\subsection{Different Attention Masks} \label{subsec:distance_mask}

\begin{figure*}[t!]
     \centering
     \begin{subfigure}[b]{0.2\textwidth}
         \centering
         \includegraphics[width=\textwidth]{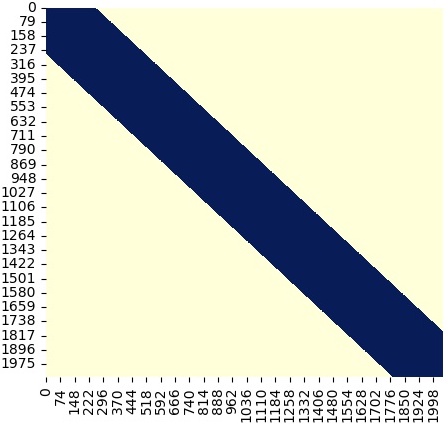}
         \caption{\slide}
         \label{fig:sliding}
     \end{subfigure}
     \hfill
     \begin{subfigure}[b]{0.2\textwidth}
         \centering
         \includegraphics[width=\textwidth]{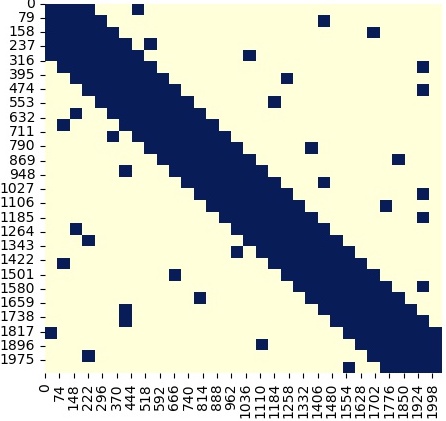}
         \caption{\bigbird}
         \label{fig:bigbird}
     \end{subfigure}
     \hfill     
     \begin{subfigure}[b]{0.2\textwidth}
         \centering
         \includegraphics[width=\textwidth]{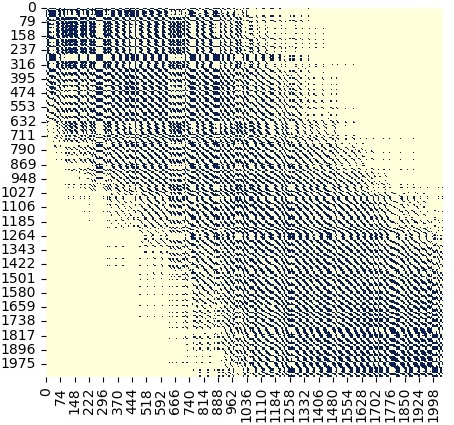}
         \caption{\distance}
         \label{fig:distance}
     \end{subfigure}
     \hfill
     \begin{subfigure}[b]{0.29\textwidth}
         \centering
         \includegraphics[width=\textwidth]{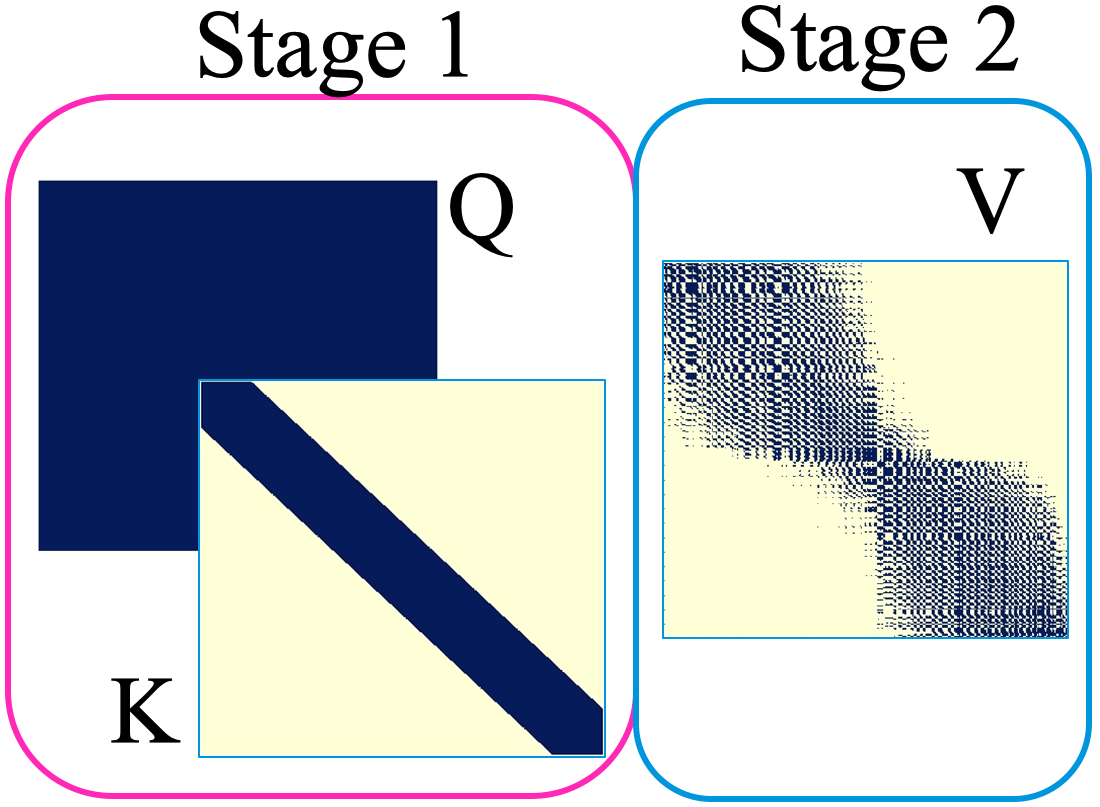}
         \caption{\hybrid}
         \label{fig:hybrid}
     \end{subfigure}
        \caption{Visualization of our models' different types of attention mask for real samples from RVL-CDIP dataset~\cite{harley2015icdar} with limit length of 2048 and context size 512 (for both textual and spatial cases). 
        Fig~\ref{fig:sliding} is sliding window (\slide),
        Fig~\ref{fig:bigbird} is sliding window in blocks with 1-per-block random blocks (\bigbird), 
        Fig~\ref{fig:distance} is spatial-based distance mask, and 
        Fig~\ref{fig:hybrid} is the combination of sliding window and distance modes.
        \textit{Legend}: Attention mask may only have values of 0 and 1, which are represented as the light-yellow background and dark-blue foreground colors, respectively. 
        }
        \label{fig:distance_mask}
\end{figure*}



We are motivated by the fact that in rich documents with multimodal contents, the relationship of words not only follows the consecutive, sequential nature of texts but also in the boxes or sections organized in many complicated forms, in which spatial input offers essential information in addition to text.  Furthermore, we argue that in dealing with long documents, we should not put extra overhead on the already-heavy transformer-based models in both computation and memory perspectives. As a result, we employ neither additional embedding techniques nor complicated encoding or fusing methods as in many other approaches (see Section~\ref{sec:related} for more information), and instead focus on making the attention, the main cost of those models, lightweight and effective by having a shortened yet flexible context information of textual and/or spatial input.

In the following, we begin to describe the original transformer attention mechanism and the different approaches that we propose specifically for long multimodal documents, using 1D and 2D input data.

\textbf{Original Attention Masks} \ For the original transformer-based architectures~\cite{vaswani2017attention}, 
in each of their layers, the attention score is calculated by two main steps, as formulated in Equations (\ref{eq:one}) and (\ref{eq:two}),

\begin{align}
     \text{score}(\mathbf{Q}, \mathbf{K}) &= \text{softmax}\left(\frac{\mathbf{Q}\mathbf{K}^T}{\sqrt{d_k}}\right)  
     \label{eq:one} 
\\
    \text{attn\_score}(\mathbf{Q}, \mathbf{K}, \mathbf{V}) &= \text{score}(\mathbf{Q}, \mathbf{K}) \cdot \mathbf{V} \label{eq:two} 
\end{align}
\normalsize 

\noindent where $\mathbf{Q}, \mathbf{K}, \mathbf{V}$ stand for the learnable Query, Key, and Value matrices respectively. Given the lengths of these three matrices are all $N$, which is also the input length, the complexity of each step is $\mathcal{O}(N^2)$. 

We usually refer to this attention mechanism as full attention because each single input token attends to all $N$ available tokens including itself, which makes it impractical in terms of both computation and memory in the cases of long documents. As a result, proper changes have to be made as described below in our proposed attention approaches. 

\textbf{Sliding-Window Masks} (Figure~\ref{fig:sliding}) \
We use the sliding-window approach as inspired from~\citet{beltagy2020longformer}, which limits the context for each token from $N$ down to a smaller $M$, e.g. $N=4096, M=512$, and so the complexity is essentially reduced to  $\mathcal{O}\left(NM\right)$.


\begin{align}
    \mathbf{K}_w &= \text{get\_window}(\mathbf{K}) \label{eq:three} 
\\
    \text{score}(\mathbf{Q}, \mathbf{K}) &= \text{softmax}\left(\frac{\mathbf{Q}\mathbf{K}_w^T}{\sqrt{d_k}}\right) \label{eq:four} 
\\
    \mathbf{V}_w &= \text{get\_window}(\mathbf{V}) \label{eq:five} 
\\
    \text{attn\_score}(\mathbf{Q}, \mathbf{K}, \mathbf{V}) &= \text{score}(\mathbf{Q}, \mathbf{K}) \cdot \mathbf{V}_w \label{eq:six} 
\end{align}

\normalsize
Using that intuition, the calculations are now changed to Equations (\ref{eq:three}--\ref{eq:six}), with the added \texttt{get\_window} steps in Equations (\ref{eq:three}) and (\ref{eq:five}). This change is simplistic because while significantly reducing the heavy blueprint of the full attention, it retains a consistent pattern of token arrangement for fast implementation~\footnote{{To enable fast calculations in Equations (\ref{eq:four}) and (\ref{eq:six}) with now-changed matrix shapes, one has to extract and chunk the context for all tokens in a way that can exploit fast matrix multiplication (e.g. by using \texttt{einsum})}}.

\textbf{Sliding-Window plus Random Token Masks} (Figure~\ref{fig:bigbird}) \
On top of sliding windows, we add a few random tokens to establish more connections to the attention, as done similarly by ~\citet{zaheer2020big}. This operation essentially makes changes only to Equations~\ref{eq:three} and \ref{eq:five}, and replace them with Equations~\ref{eq:bigbird1} and \ref{eq:bigbird2}, respectively. 

\begin{align}
    \mathbf{K}_w &= \text{get\_sliding\_and\_rand\_window}(\mathbf{K}) \label{eq:bigbird1} 
    \\
    \mathbf{V}_w &= \text{get\_sliding\_and\_rand\_window}(\mathbf{V}) \label{eq:bigbird2} 
\end{align}

In more detail, the sliding-window contexts are enhanced by some random contexts added. While certainly being an extra overhead, the number of those random connections is limited to only a few, maintaining the practicality of the model in the face of long documents~\footnote{ 
Due to the introduction of those random tokens, the consistent pattern of sliding windows and hence their fast implementation are largely affected. We divide the original sequence length into blocks (e.g. 512 to 8 equal blocks of length 64), to facilitate grouping and chunking, as well as to lessen the computational steps (have much fewer sliding windows) and only use 3 blocks, by default, for random connections}. 

\begin{align}
    \mathbf{K}_w &= \text{get\_2D\_spatial\_window}(\mathbf{K}) \label{eq:knn1} 
    \\
    \mathbf{V}_w &= \text{get\_2D\_spatial\_window}(\mathbf{V}) \label{eq:knn2} 
\end{align}

\textbf{Spatial Distance Masks} (Figure~\ref{fig:distance}) \
Different from previous attention types, the $M$ contextual neighbors of each token are decided by spatial (2D) information instead of textual (1D) information. In the final result, however, the spatial attention mask has the same shape as sliding windows (if they both have the same number of contextual neighboring tokens). This process comprises a couple of steps.  

First, we calculate the centers of all bounding boxes. Second, we fit the kNN algorithm to the sequence of those points based on L2 distance, resulting in a 2D distance matrix, in which each token now spatially attends to $M$ neighboring tokens. In summary, we replace Equations~\ref{eq:three} and \ref{eq:five} with Equations~\ref{eq:knn1} and \ref{eq:knn2}. 
The resulting masks consequently have a non-consecutive neighboring relationship, unlike in the traditional text-based contexts. More illustrations of those distance-based masks for real examples are also shown in Figure~\ref{fig:addl_distance_mask} in Appendix~\ref{appx:additional_mask}. 
And because of its importance, in the following, we detail the implementation notes for those new attention masks.  

\textbf{Implementation of Distance Masks} \
In terms of efficient implementation, there are certain considerations to enable the practical use of those newly proposed distance masks, which consume more computation and memory cost compared to the normal sliding window mechanism. 

First, identifying spatial neighbors for each token usually takes quadratic time, which is a great deterrent to our solution. So we choose to use \texttt{scikit-learn}'s kNN library~\footnote{{\url{https://scikit-learn.org/stable/modules/generated/sklearn.neighbors.KNeighborsClassifier.html}}} for its well-regarded efficiency and speed. 

Second, "where to create distance masks: in dataset loader or in model computation" is a key problem. We choose to create distance masks in the dataset loader for the following reasons. On one hand, the main obstacle to applying long-document attention methods is that the transformer-based models are inherently heavy. If placing the quadratic computation of those distance masks in the main model phase, the model will be significantly slower (in proportion to document lengths) and the risk of out-of-memory will be much higher (given the limitation of GPU memory). On the other hand, by preemptively computing the distance mask in the dataset loader, e.g. using Pytorch Dataloader~\footnote{{\url{https://pytorch.org/docs/stable/\_modules/torch/utils/data/dataloader.html\#DataLoader}}} and exploiting its data buffering mechanism, the data loading will not be slower by running multiple loader processes simultaneously. 

Finally, for the sliding-window attention, we inherit the implementation from Huggingface~\footnote{{\url{https://huggingface.co/transformers/model_doc/longformer.html}}}, then implement our distance-based solution on top of it. 




\subsection{Pretrain Model Variants} 

We build out MLM pretraining architecture with various attention mechanisms for long documents as described in Section~\ref{subsec:distance_mask} and compare their performances in several tasks. Since this change is only made directly to the attention, our method can be used off-the-shelf for transformer-based architecture with multimodal input.

\textbf{\slide\ Model} \ This model directly uses Sliding-Window (SW) masks for attention, which significantly reduces the computation and was shown to be effective for long documents in text-based tasks (Figure~\ref{fig:sliding}). 

\textbf{\bigbird\ Model} \ This model uses blocked Sliding-Window plus some random blocks on top (Figure~\ref{fig:bigbird}).  

\textbf{\distance\ Model} \
This model uses Spatial Distance Masks, with all neighboring contexts being preemptively computed using kNN, and is implemented in the data loading instead of transformer encoding phase, not to slow down the main process (Figure~\ref{fig:distance}).

\textbf{\hybrid\ Model}. 
In this model, we combine the spatial and textual attention masks together in a single attention pass. In detail, it is done via two steps, as shown in Equations (\ref{eq:three}--\ref{eq:six}), with Equation (\ref{eq:three}) now being replaced by Equation (\ref{eq:knn1}). This is a possible adjustment since these two steps are separated and both preserve the logic and shapes of matrices in their calculation. Our motivation and intuition are to combine the benefits of both textual and spatial information in a single attention pass (Figure~\ref{fig:hybrid}). 

\section{Experiments} \label{sec:experiments} 

In this section, we describe our experimental methodology to evaluate our proposed approach of flexible attention using different contextual input information.

\subsection{Tasks and Datasets} \label{subsec:dataset} 
 

 
 \textbf{Pretraining} \ We use \textbf{IIT-CDIP Test Collection 1.0}~\footnote{{\url{https://ir.nist.gov/cdip/}}} dataset for our MLM pretraining task.  This is a large-scale dataset with over 6M multi-page documents and around 11M pages in total (each page is stored as a scanned image and is preprocessed by an OCR engine). 
 
 
 \textbf{Document Classification} \ 
 This document classification task uses \textbf{RVL-CDIP}~\cite{harley2015icdar} dataset, which is a subset of the pretraining dataset IIT-CDIP. 
 It comprises 16 classes and each class equally has 25K grayscale images. 
 All of these 400K images in combination are split into 320K images for training and 40K images each for validation and testing. 
For more statistics on this dataset, the document length distribution is shown in Figure~\ref{fig:ldistribution}. 
 
 
 \textbf{Sequence Labeling} \  There are two datasets for this task, namely Kleister-NDA and FunSD.

 \textbf{1) FunSD}~\cite{jaume2019}\footnote{{\url{https://guillaumejaume.github.io/FUNSD}}} \ 
 This is a lightweight dataset that has 199 noisy scanned forms, which contain around 31K words and 9.7K entities with 7 given token classes. 
Although it is not a long-document dataset (all documents have < 512 words), it is a popular dataset used by many document intelligence models and is also useful for ablation studies on how long document models perform on a short document dataset, as we show in Section~\ref{subsec:short}. 
  
  \textbf{2) Kleister-NDA}~\cite{gralinski2020kleister,stanislawek2021kleister}\footnote{{\url{https://github.com/applicaai/kleister-nda}}} \ 
This dataset has 540 documents in total (254 train, 83 validation, and 203 test) with 2,160 entities annotated and an average of 2,540 words per document.
Due to the difficulty in reproducibility with unclear results post-processing, this task is cast similarly to FunSD with 4 classes. Consequently, we report the evaluation results of our models along with all other methods' reproduced outcomes using the same preprocessing steps and metrics, in order to maintain fair comparisons. 



\subsection{Baselines}

We pretrain our 4 model variants (Figure~\ref{fig:distance_mask}) with the MLM objective and then compare them with the following baseline groups: 

     \textbf{Text}: This group consists of models that only accept text input including BERT~\cite{devlin2018bert}, RoBERTa~\cite{liu2019roberta}, and other long models including Bigbird~\cite{zaheer2020big} and Longformer~\cite{beltagy2020longformer}~\footnote{{Our \slide \ and \bigbird \ models share the similarity with those last two ones, with the difference of handling multimodal input for document intelligence.}}.  
     
     \textbf{Text+Layout}: This group contains models that accept both text and layout information, including LayoutLM~\cite{xu2020layoutlm} variants. 


\subsection{Results and Discussions}
\label{subsec:results}

\begin{table}[t]
\fontsize{11}{11}\selectfont
\setlength\tabcolsep{3.0pt}
\centering
\footnotesize
\begin{tabular}{clcc}
\Xhline{3\arrayrulewidth}
\multicolumn{1}{l}{Type} & Model & SeqLen & Acc (\%) \textcolor{gg}{$\uparrow$} \\ \hline
\multirow{8}{*}{Text} 
     & BERT-base & 512 &  $89.81$ \\
     & RoBERTa-base & 512 & $90.06$ \\
     & BERT-large & 512 & $89.92$ \\
     & RoBERTa-large & 512 & $90.11$ \\ 
\cline{2-4}
     & Bigbird-base  & 4096 & $93.48$ \\ 
     & Longformer-base  & 4096 & $93.85$ \\ 
     & Bigbird-large  & 4096 & $93.34$ \\ 
     & Longformer-large  & 4096 & $93.73$ \\
\hline
\multirow{6}{*}{Text+Layout} 
     & LayoutLM-base & 512 & $91.88$ \\
     & LayoutLM-large & 512 & $91.90$ \\
\cline{2-4}
     & Ours \slide  & 4096 & $94.50$ \\
     & Ours \bigbird  & 4096 & $\textbf{95.25}$ \\
     & Ours \distance \  & 4096 & $94.79$ \\
     & Ours \hybrid \  & 4096 &  $94.69$ \\
\Xhline{3\arrayrulewidth}
\end{tabular}
\caption{
\label{tab:rvl_cdip} 
Classification accuracy for RVL-CDIP. For this long-document dataset, the models capable of using 4096 words uniformly beat other models and layout information helps with the task compared with using Text input. All our long models show their advantages on this long dataset.
}
\end{table}

\textbf{Document Classification} \ As shown in Table~\ref{tab:rvl_cdip}, long models (SeqLen\footnote{SeqLen is short for Sequence Length} 4096) clearly outperform short ones in both baseline groups, with or without layout information added to the input. Furthermore, all our 4 model variants outperform all the baselines. 

This result concurs with our observation that long documents have valuable information spanned across the length. And importantly, our models show advantages of handling long multimodal input, and hence are more practical with real data that are usually longer than 512 tokens. 



\begin{table}[t]
\fontsize{11}{11}\selectfont
\setlength\tabcolsep{3.0pt}
\centering
\footnotesize
\begin{tabular}{clcc}
\Xhline{3\arrayrulewidth}
\multicolumn{1}{l}{Type} & Model & SeqLen & F1  \textcolor{gg}{$\uparrow$} \\ \hline
\multirow{4}{*}{Text} 
     & BERT-base & 512 & $47.06$ \\
     & BERT-large & 512 & $52.66$ \\
\cline{2-4}
     & Longformer-base & 4096 & $61.78$ \\ 
     & Bigbird-base  & 4096 & $46.98$ \\ 
\hline
\multirow{6}{*}{Text+Layout} 
     & LayoutLM-base & 512 & $55.69$ \\ 
     & LayoutLM-large &512 & $61.95$ \\ 
\cline{2-4}
     & Ours \slide  & 4096 & $\textbf{64.06}$ \\ 
     & Ours \bigbird & 4096 &$58.92$ \\ 
     & Ours \distance & 4096 & $57.01$ \\ 
     & Ours \hybrid \   & 4096 &  $44.70$ \\ 
\Xhline{3\arrayrulewidth}
\end{tabular}
\caption{
\label{tab:nda} 
Results on Kleister-NDA. Although this dataset is challenging, long models still show advantages over short ones. 
}
\end{table}

\textbf{Sequence Labeling with Kleister-NDA}\footnote{{The results are from the validation split due to no annotation for the test split provided in the dataset.}} \
Comparing the ``base'' versions (separated from their ``large'' counterparts), Table~\ref{tab:nda} shows that most of our models, which are also the ``base'' ones, clearly have better scores. Particularly, our \slide \ model is the best performer. 

Furthermore, our \hybrid \ is not performing equally well. Our hypothesis is that the OCR engine cannot understand the decoying annotation in this dataset, and thus generates spatial results that do not correlate well with the text. Consequently, the combination of textual and spatial information does not result in the benefits of those two.

\subsection{Ablation: Long Models on Short Dataset} \label{subsec:short} 

The purpose of this study is to explore how long models perform on short documents, which also appear in practice, to see whether they can generalize their performance to shorter data. 

\begin{table}[h]
\fontsize{11}{11}\selectfont
\setlength\tabcolsep{3.0pt}
\centering
\footnotesize
\begin{tabular}{clcc}
\Xhline{3\arrayrulewidth}
\multicolumn{1}{l}{Type} & Model & SeqLen & F1 \textcolor{gg}{$\uparrow$} \\ \hline
\multirow{8}{*}{Text} 
     & BERT-base & 512 & $60.3$ \\
     & RoBERTa-base & 512 &  $66.5$ \\
     & BERT-large & 512 &  $65.6$ \\
     & RoBERTa-large & 512 &  $70.7$ \\ 
\cline{2-4}     
     & Bigbird-base  & 4096 &  $45.8$ \\ 
     & Longformer-base  & 4096 &  $71.4$ \\ 
     & Bigbird-large  & 4096 &  $46.8$ \\ 
     & Longformer-large  & 4096 &  $73.5$ \\
\hline
\multirow{8}{*}{Text+Layout} 
     & LayoutLM-base & 512 &  $78.7$ \\
     & LayoutLM-large & 512 &  $\textbf{79.0}$ \\
\cline{2-4}
     & Ours \slide  & 4096 &  $69.9$ \\
     & Ours \bigbird  & 4096 & $77.1$ \\
     & Ours \distance \  & 4096 & $64.0$ \\
     & Ours \hybrid \  & 4096 &  $61.8$ \\
\Xhline{3\arrayrulewidth}
\end{tabular}
\caption{
\label{tab:funsd} 
Results on FunSD dataset. As usual, layout information is helpful in boosting performance. However, long models do not perform well compared with short models on this small, short-document dataset. 
}
\end{table}

Table~\ref{tab:funsd} shows that on FunSD, we see again that layout information generally helps in the case of multimodal input.
However, long models do not perform well compared to short ones, although the gap between the best of ours and the baselines are not very far away (77.1 vs. 79.0). 
The main reason is that long models essentially have much more parameters than short ones. And not only is FunSD short, it is also very small. As a result, the limited phase of fine-tuning on only 199 samples can hardly tune parameters well for good results. 
Especially, since all documents are short, most long input to the model is zero padding and thus not enough for contributing for better scores. 

Another reason is that long models have their embedding representations trained for the length of 4096 tokens and hence are hard to adapt to 512-token input with just a few fine-tuning steps. As a result, analyzing the data well to design suitable pretraining and fine-tuning models is very important. 


The next 2 studies will explore the implications of the newly-added spatial attention masks in our models.

\subsection{Ablation: Different-Length Documents}

This study aims to explore how the models work if we do not cut any information from documents (the models take input up to their maximum length limit). 
Out of 40K test samples in RVL-CDIP, there are 9268 samples with length $\geq 512$, 2312 with length $\geq 1024$, and only 106 with length $\geq 2048$. 

\begin{figure}[t!]
    \centering
  \includegraphics[width=0.4\textwidth]{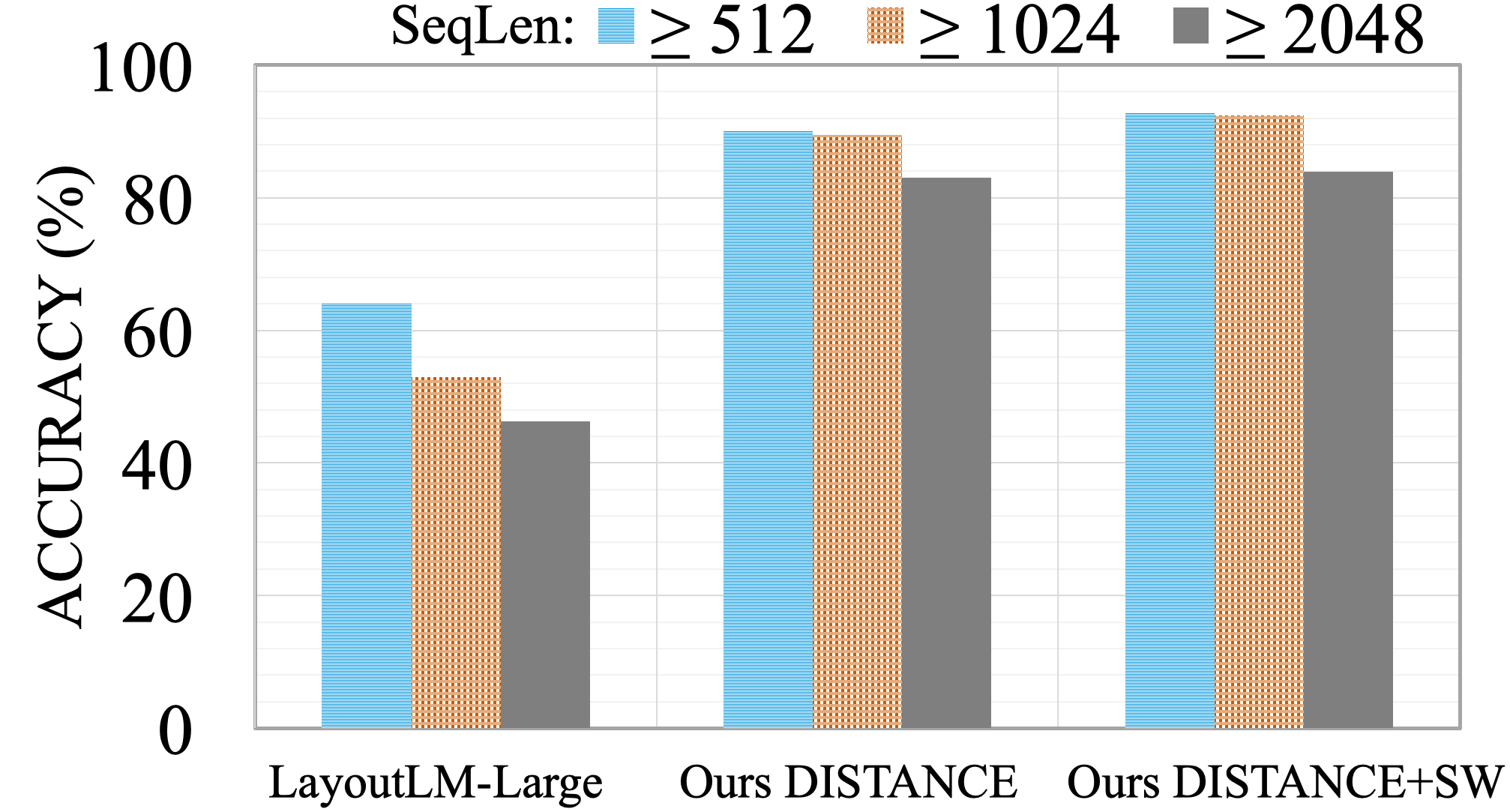}
  \caption{RVL-CDIP performance on different document types based on their original lengths (i.e. without purging) with LayoutLM (with the best ``large'' version) and our spatial models (\distance \ and \hybrid). Our models are consistently better. 
  }
    \label{fig:diff_doc_types}
\end{figure}

Figure~\ref{fig:diff_doc_types} shows the consistent observation that our models are much better than LayoutLM, and yet perform slightly worse as the original document length increases.
There could be several possible reasons for this behavior: the models are not well pretrained and/or fine-tuned, many long documents have lots of confusing parts, or there are many noises in OCR results.

\subsection{Ablation: Different Max Input Lengths} \label{subsec:diff_len} 

Given the pretrained models that can accept input up to 4096 tokens, we finetune them with the input of different maximum lengths, i.e. excess will be purged. 
As a result, we use all 40K test samples in RVL-CDIP for this study. 

\begin{figure}[t!]
    \centering
  \includegraphics[width=0.4\textwidth]{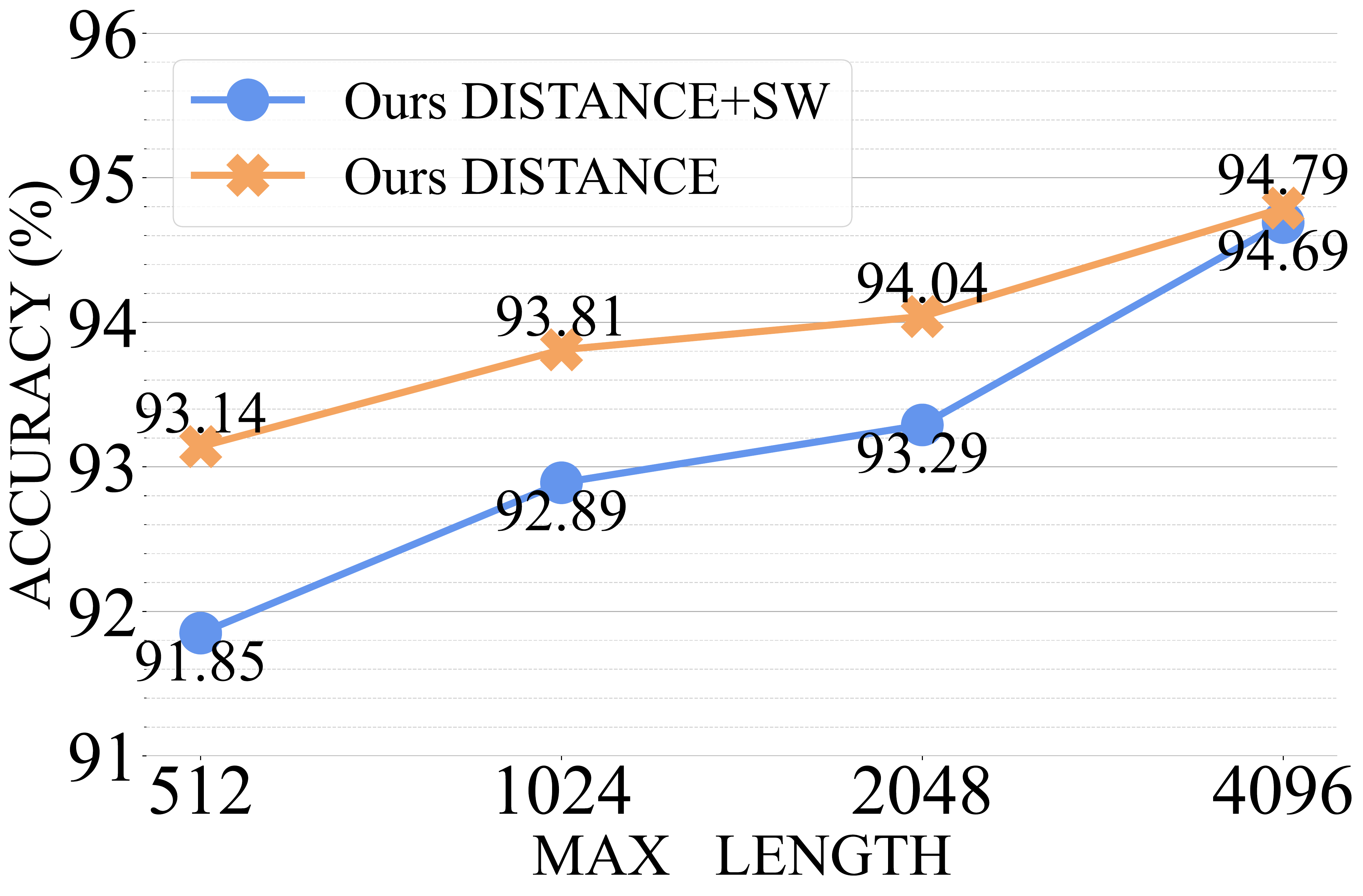}
  \caption{RVL-CDIP performance on different maximum lengths using our \distance \ and \hybrid \ models. For each of lengths 512, 1024, 2048, and 4096, the test set contains the same 40K samples. A longer maximum length gives better results.}
    \label{fig:diff_len}
\end{figure}

As shown in Figure~\ref{fig:diff_len}, our models are better and better as more tokens are absorbed, thus once again confirming our intuition that valuable information is spanned across the length. As a result, if the model capacity permits, we should not limit the capacity to 512 tokens as in most current models in the literature.

\subsection{Further Discussion on Spatial Masks} 
As seen in the above experimental results, direct usage of 2D layout context information in the transformer attention has some advantages. However, its performance does not match the typical usage of 1D textual information. This might be discouraging at first since introducing spatial masks brings heavier computation compared to textual ones. We hypothesize the drawbacks are due to some objective limitations. First, the kNN suffers some inaccuracy compared with normal (and slow) calculations. Second, the performance of the whole pipeline heavily depends on OCR quality, e.g. in Kleister-NDA with decoy design, OCR results are not well aligned with the text. Consequently, we conjecture that with future development in OCR technologies, the use of spatial masks would be more and more helpful in practice. 

\section{Conclusion and Discussion} 

We propose a versatile solution for long document understanding, in which the shortened context can be used in the form of textual and/or layout input for the attention mechanism in a flexibly pluggable manner. We keep our approach simple by not putting extra overhead on complicated embedding or encoding methods. Despite its simplicity, our solution has shown promising experimental results on document understanding tasks with long, multimodal input. 
In the future, we will further reduce the memory consumption of models with given multimodal input and speed up the pretraining. 
Similar to LayoutLM, pretraining usually takes 80 hrs/epoch with 8 V100 GPUs. Thus there are certainly lots of room for improvement to make these models more efficient and practical.


\bibliography{mybib}
\bibliographystyle{acl_natbib}

\newpage\phantom{blabla}
\newpage

\appendix

\section{More Information on the Pretrain Task}
\label{appx:pretrain}

\textbf{Pretrain Data Preprocessing} \
As described, for pretrain model we retain the same OCR engine for generating and aligning layout and text information from LayoutLM~\cite{xu2020layoutlm}. The task is also the same, which is Masked Language Modeling (MLM). 
To deal with long documents, we have to implement the additional sliding-window, random-block and distance-based masks. 

\textbf{Pretrain Model Implementation} \ 
Our solution only makes changes to the attention module, in which uses can choose to use any types of attention masks from the 4 variants illustrated in Figure~\ref{fig:distance_mask}. 

For the \slide \ and \bigbird \ models which are also our new pretrain models, we implement the layout-related part on top of the original BigBird~\footnote{\url{https://huggingface.co/transformers/model_doc/bigbird.html}} and Longformer~\footnote{{\url{https://huggingface.co/transformers/model_doc/longformer.html}}} implementations from Huggingface's transformers, respectively.  
Otherwise the distance-based masks, which are employed in \distance \ and \hybrid \ models, are newly implemented as a pluggable module.

\begin{table}[t]
\fontsize{11}{11}\selectfont
\setlength\tabcolsep{3.0pt}
\centering
\footnotesize
\begin{tabular}{lc}
\Xhline{3\arrayrulewidth}

Parameter Name & Value   \\
\hline
do\_lower\_case & true \\ 
fp16 & true \\ 
fp16\_backend & amp \\ 
gradient\_accumulation\_steps & 4 \\ 
max\_seq\_length & 4096 \\ 
max\_2d\_position\_embeddings & 1024 \\ 
max\_steps & 1000000 \\ 
model\_name\_or\_path & allenai/longformer-base-4096 \\ 
dataloader\_num\_workers & 64 \\ 
tasks & mask\_lm \\ 
optimizer & transformers\_AdamW \\ 
learning\_rate & 5e-5 \\ 
warmup\_ratio & 0.1 \\ 
weight\_decay & 0.01 \\ 
whole\_word\_masking & false \\ 
add\_prefix\_space & true \\ 
attention\_window & 512 \\

\Xhline{3\arrayrulewidth}
\end{tabular}
\caption{
\label{tab:pretrain} 
Main pretrain hyperpameters on the MLM pretraining task for the ITT-CDIP large-scale dataset. There are 3 variants share this set of parameters that are Ours \slide, Ours \distance \ and Ours \hybrid \ models. All of them use the pretrained weights from Longformer-base~\cite{beltagy2020longformer} models. 
}
\end{table}

\textbf{Training MLM} \ 
We pre-train the task on the IIT-CDIP datasets, using a single-node multi-GPU mode. Each job was run on a server with 8 V100 Nvidia GPUs, each of which has 32GB of memory and fast processors. For text-only models, please refer to LayoutLM's github~\footnote{{\url{https://github.com/microsoft/unilm}}}. 

For \slide \ model, we use the public pretrained weights from Lomgformer~\cite{beltagy2020longformer}. Other of our models employ the same set of parameters, except for the pretrained weights, in which \bigbird \ model uses the weights from Bigbird~\cite{zaheer2020big} and the last two models having distance masks (\distance \ and \hybrid \ models) use the same pretrained weights as \slide \ model, as demonstrated in Table~\ref{tab:pretrain}. 

It is also worth noting that the pretrained weights from Longformer and Bigbird models are useful even for the models using distance masks because those two model families support documents with length 4096, so the position embeddings are helpful. For speed and memory tradeoff, we limit the context for distance masks to only 128 (vs. 512 in textual contexts), without sacrificing much performance, as reported in Section~\ref{subsec:results}. 

\textbf{Training Notes} \ 
Although not reported in the main content, we note some lessons learned from the pretraining task. 
As we observe, the Ours \slide \ model consistently achieves the best results, while consuming the least GPU memory. For the base model, it only consumes about 7 GB GPU memory and Ours \hybrid \ that uses sliding-window attention on its half processing also consumes about 9 GB memory. Both models, as a result, can be deployed well on a broad range of GPUs in the market. 

Unlike those conveniences, Ours \bigbird \ and Ours \distance \ do not share the same advantages. In fact, they consumes about more than 30GB GPU memory each, limiting their practicality. 
We hypothesize the main reason for such drawbacks is that they have random, inconsistent patterns, and hence there is no efficient way to take advantage of fast memory-efficient and fast matrix operations. 

Finally, although showing promising practical behaviors, all baselines and our models, and almost any transformer-based ones are certainly not lightweight models. And although there are advancements in compressing those heavy models (e.g. \cite{touvron2021training,frankle2018lottery}, there seems to be a considerable way to go for making these model run on mobile devices in the near future.

\section{More information on Finetuning Tasks} 

As described in the main content, after pretraining, the saved models are the backbone for the respective fine-tuning model types. For that reason, the parameters are mostly shared with their pretrain counter-part models, e.g. Table~\ref{tab:pretrain} for Ours \slide \ models. Generally, we keep the same optimizer and batch size of 32 (combined across all used parallel GPUs). 

For \textbf{RVL-CDIP} in the document classification task, we use the \texttt{SequenceClassification} model type. On top of the pretrain skeleton, we add a small classifier with 2 fully-connected layers and a drop-out layer in between.  The final output is the single class for the whole sequence/document.  

For \textbf{FunSD} and \textbf{Kleister-NDA} datasets, we instead use the \texttt{TokenClassification} model type, which is designed to classify all-document entities. The similar classifier is added to the pretrained skeleton, now with a different usage in which each token/entity is to be classified into one of the number of given classes. 

What's more, to preprocess these two datasets, we have to ingest all available document tokens. Likewise, with documents longer than the maximum lengths, we need to cut those documents, and recursively treat the overflowing parts in the same way. In terms of implementation, unlike FunSD that is lightweight, we always want to avoid loading the whole dataset into the memory but rather taking advantage of the data buffering in feeding to the models. As a result, we pre-process all data first, save them to disks and only load the respective parts when needed.

\begin{figure}[ht!]
     \centering
     \begin{subfigure}[b]{0.2\textwidth}
         \centering
         \includegraphics[width=\textwidth]{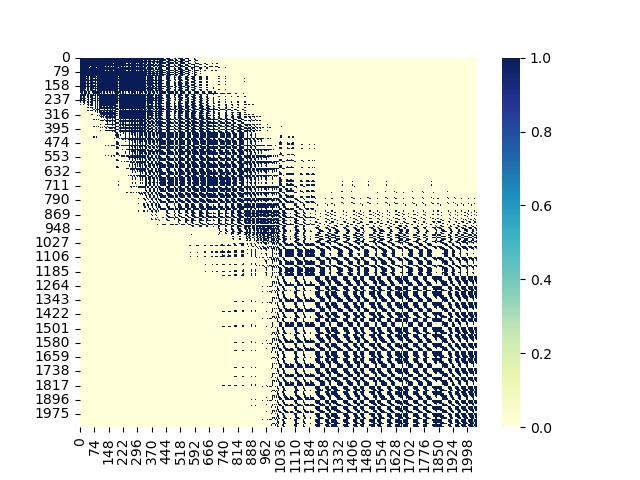}
     \end{subfigure}
     \begin{subfigure}[b]{0.2\textwidth}
         \centering
         \includegraphics[width=\textwidth]{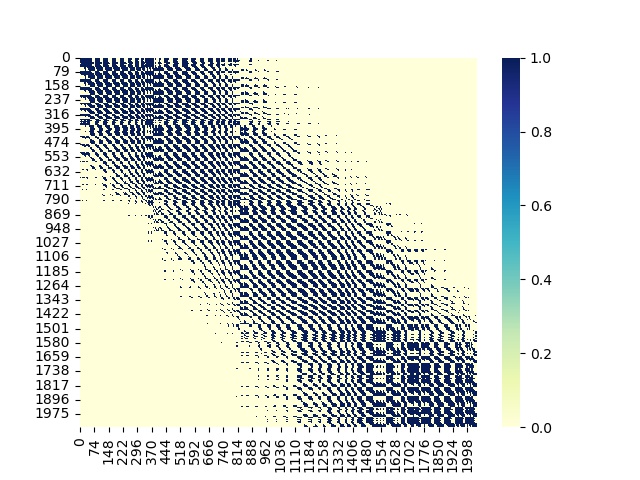}
     \end{subfigure}
\\
     \begin{subfigure}[b]{0.2\textwidth}
         \centering
         \includegraphics[width=\textwidth]{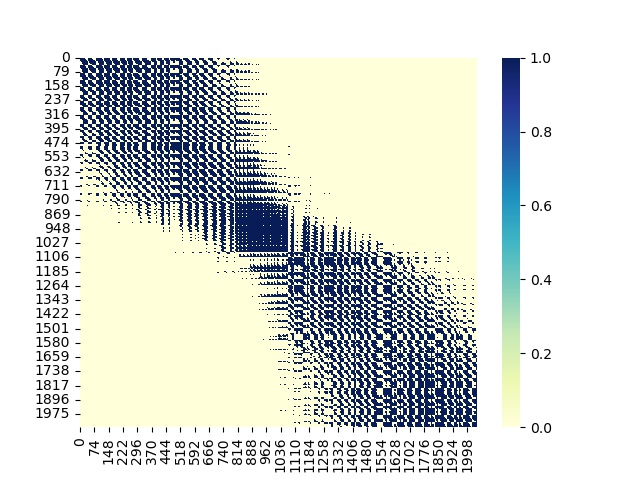}
     \end{subfigure}
     \begin{subfigure}[b]{0.2\textwidth}
         \centering
         \includegraphics[width=\textwidth]{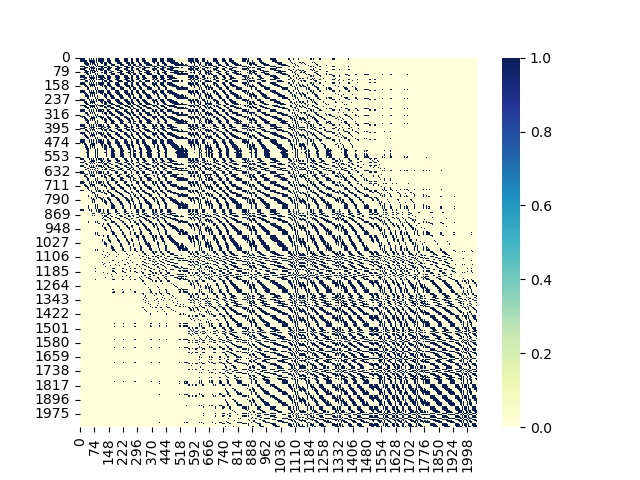}
     \end{subfigure}
\\
     \begin{subfigure}[b]{0.2\textwidth}
         \centering
         \includegraphics[width=\textwidth]{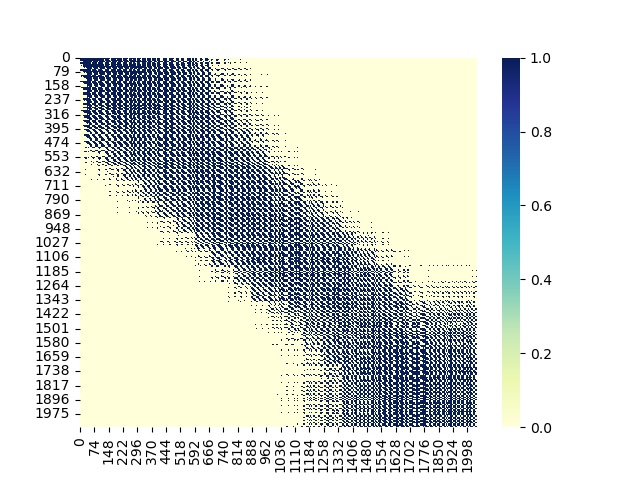}
     \end{subfigure}
     \begin{subfigure}[b]{0.2\textwidth}
         \centering
         \includegraphics[width=\textwidth]{figs/example_12.npy_knn.jpeg}
     \end{subfigure}
\\
     \begin{subfigure}[b]{0.2\textwidth}
         \centering
         \includegraphics[width=\textwidth]{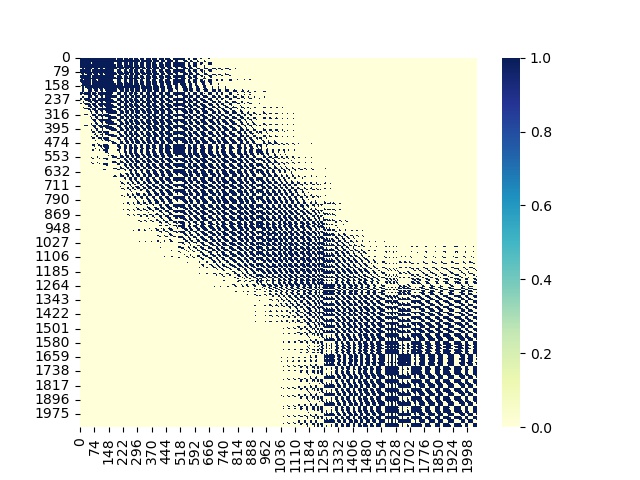}
     \end{subfigure}
     \begin{subfigure}[b]{0.2\textwidth}
         \centering
         \includegraphics[width=\textwidth]{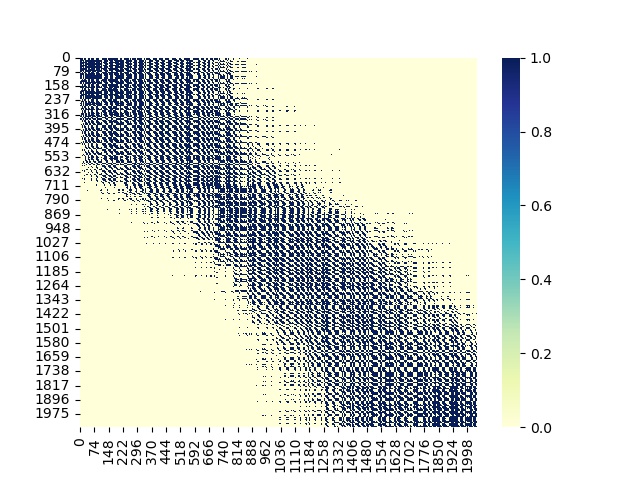}
     \end{subfigure}
\\
     \begin{subfigure}[b]{0.2\textwidth}
         \centering
         \includegraphics[width=\textwidth]{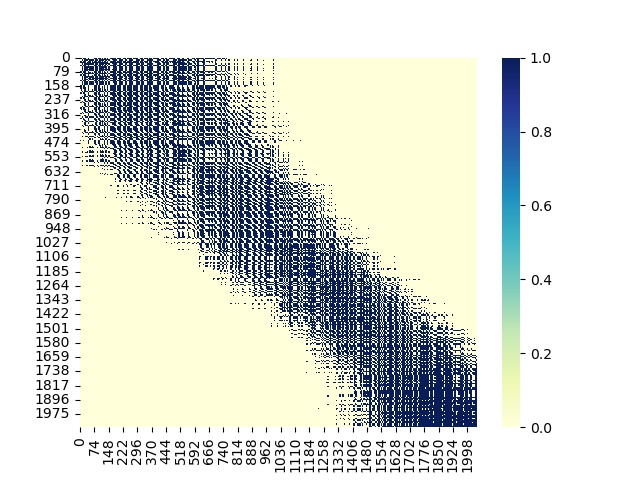}
     \end{subfigure}
     \begin{subfigure}[b]{0.2\textwidth}
         \centering
         \includegraphics[width=\textwidth]{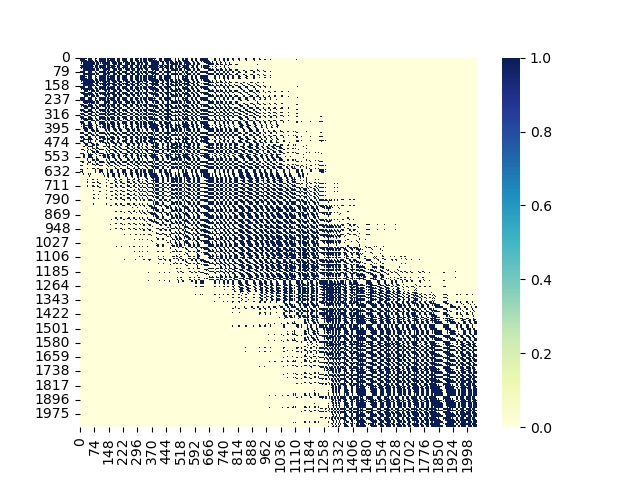}
     \end{subfigure}
\\
     \begin{subfigure}[b]{0.2\textwidth}
         \centering
         \includegraphics[width=\textwidth]{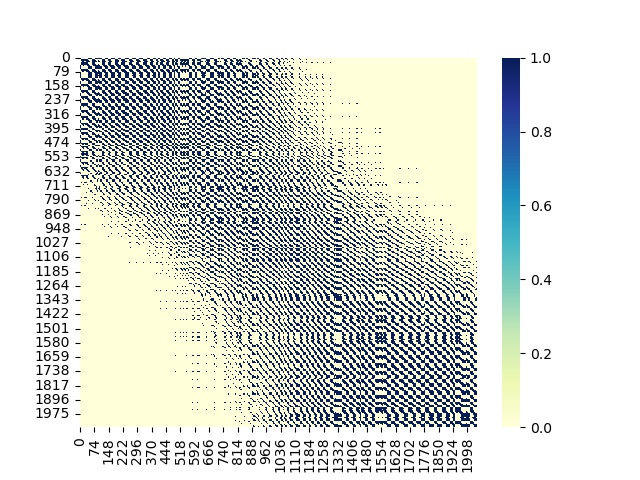}
     \end{subfigure}
     \begin{subfigure}[b]{0.2\textwidth}
         \centering
         \includegraphics[width=\textwidth]{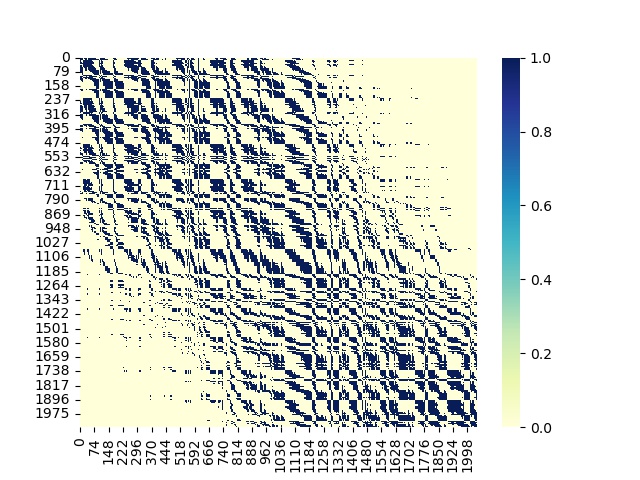}
     \end{subfigure}
     
        \caption{More distance masks from RVL-CDIP samples with the limit length of 2048 and 512 neighbors each. 
        }
        \label{fig:addl_distance_mask}
\end{figure}

\textbf{Additional Information for Kleister-NDA} It is worth recalling that the evaluation of it is tricky if using the provided official GEval evaluation script~\cite{gralinski2020kleister}\footnote{{\url{https://github.com/applicaai/kleister-nda}}}. In detail, given the predited tokens, one has to retrieve the associated texts in a group. For example, the beginning of an entity group usually starts with a class beginning with "B-", followed by a series of "I-" tokens. However, there is no guarantee that the prediction will always return a group having this meaningful pattern, let alone many other complicated cases that can happen. Such complications make the post-processing of the prediction-- before feeding to GEval--very difficult and importantly, not easily reproducible. In fact, amongst recent papers that report performance on this dataset (e.g. in ~\citet{xu2020layoutlmv2,appalaraju2021docformer}), there is reference code with which for us to compare. 

Consequently, we treat this dataset the same as FunSD, given their similarity in annotation. In addition, because this dataset is larger and much more difficult (due to decoying texts) compared to FunSD, we analyze the train dataset and employ the weighted loss based on the distribution the given labels. As a result, our method is more transparent and reproducible. 

\section{Additional Samples on Distance Masks} \label{appx:additional_mask}
Complementary to Figures~\ref{fig:distance} and \ref{fig:hybrid}, we present more distance masks based on real samples taken from RVL-CDIP with the same setting in Figure~\ref{fig:addl_distance_mask}.

\end{document}

%% file: math_commands.tex
\usepackage{amsmath,amsfonts,bm}
\usepackage{amsthm}
\usepackage{mathtools}

        {\hspace*{\fill}$\Box$\par}

\DeclarePairedDelimiterX{\infdivx}[2]{(}{)}{%
  #1\;\delimsize\|\;#2%
}

\def\eqref#1{equation~\ref{#1}}

\def\1{\bm{1}}

\DeclareMathAlphabet{\mathsfit}{\encodingdefault}{\sfdefault}{m}{sl}
\SetMathAlphabet{\mathsfit}{bold}{\encodingdefault}{\sfdefault}{bx}{n}














%% file: main.bbl
\begin{thebibliography}{24}
\expandafter\ifx\csname natexlab\endcsname\relax\def\natexlab#1{#1}\fi

\bibitem[{Ainslie et~al.(2020)Ainslie, Ontanon, Alberti, Cvicek, Fisher, Pham,
  Ravula, Sanghai, Wang, and Yang}]{ainslie2020etc}
Joshua Ainslie, Santiago Ontanon, Chris Alberti, Vaclav Cvicek, Zachary Fisher,
  Philip Pham, Anirudh Ravula, Sumit Sanghai, Qifan Wang, and Li~Yang. 2020.
\newblock Etc: Encoding long and structured inputs in transformers.
\newblock In \emph{Proceedings of the 2020 Conference on Empirical Methods in
  Natural Language Processing (EMNLP)}, pages 268--284.

\bibitem[{Appalaraju et~al.(2021)Appalaraju, Jasani, Kota, Xie, and
  Manmatha}]{appalaraju2021docformer}
Srikar Appalaraju, Bhavan Jasani, Bhargava~Urala Kota, Yusheng Xie, and
  R~Manmatha. 2021.
\newblock Docformer: End-to-end transformer for document understanding.
\newblock \emph{ICCV}.

\bibitem[{Beltagy et~al.(2020)Beltagy, Peters, and
  Cohan}]{beltagy2020longformer}
Iz~Beltagy, Matthew~E Peters, and Arman Cohan. 2020.
\newblock Longformer: The long-document transformer.
\newblock \emph{arXiv preprint arXiv:2004.05150}.

\bibitem[{Chen et~al.(2020)Chen, Li, Yu, El~Kholy, Ahmed, Gan, Cheng, and
  Liu}]{chen2020uniter}
Yen-Chun Chen, Linjie Li, Licheng Yu, Ahmed El~Kholy, Faisal Ahmed, Zhe Gan,
  Yu~Cheng, and Jingjing Liu. 2020.
\newblock Uniter: Universal image-text representation learning.
\newblock In \emph{European Conference on Computer Vision}, pages 104--120.
  Springer.

\bibitem[{Devlin et~al.(2019)Devlin, Chang, Lee, and
  Toutanova}]{devlin2018bert}
Jacob Devlin, Ming-Wei Chang, Kenton Lee, and Kristina Toutanova. 2019.
\newblock \href {https://doi.org/10.18653/v1/N19-1423} {{BERT}: Pre-training of
  deep bidirectional transformers for language understanding}.
\newblock pages 4171--4186.

\bibitem[{Frankle and Carbin(2019)}]{frankle2018lottery}
Jonathan Frankle and Michael Carbin. 2019.
\newblock The lottery ticket hypothesis: Finding sparse, trainable neural
  networks.
\newblock \emph{ICLR}.

\bibitem[{Grali{\'n}ski et~al.(2020)Grali{\'n}ski, Stanis{\l}awek,
  Wr{\'o}blewska, Lipi{\'n}ski, Kaliska, Rosalska, Topolski, and
  Biecek}]{gralinski2020kleister}
Filip Grali{\'n}ski, Tomasz Stanis{\l}awek, Anna Wr{\'o}blewska, Dawid
  Lipi{\'n}ski, Agnieszka Kaliska, Paulina Rosalska, Bartosz Topolski, and
  Przemys{\l}aw Biecek. 2020.
\newblock Kleister: A novel task for information extraction involving long
  documents with complex layout.
\newblock \emph{arXiv preprint arXiv:2003.02356}.

\bibitem[{Guillaume~Jaume(2019)}]{jaume2019}
Jean-Philippe~Thiran Guillaume~Jaume, Hazim Kemal~Ekenel. 2019.
\newblock Funsd: A dataset for form understanding in noisy scanned documents.
\newblock In \emph{Accepted to ICDAR-OST}.

\bibitem[{Harley et~al.(2015)Harley, Ufkes, and Derpanis}]{harley2015icdar}
Adam~W Harley, Alex Ufkes, and Konstantinos~G Derpanis. 2015.
\newblock Evaluation of deep convolutional nets for document image
  classification and retrieval.
\newblock \emph{International Conference on Document Analysis and Recognition}.

\bibitem[{Hong et~al.(2022)Hong, Kim, Ji, Hwang, Nam, and Park}]{hong2021bros}
Teakgyu Hong, Donghyun Kim, Mingi Ji, Wonseok Hwang, Daehyun Nam, and Sungrae
  Park. 2022.
\newblock Bros: A pre-trained language model focusing on text and layout for
  better key information extraction from documents.
\newblock \emph{AAAI}.

\bibitem[{Huang et~al.(2022)Huang, Lv, Cui, Lu, and Wei}]{huang2022layoutlmv3}
Yupan Huang, Tengchao Lv, Lei Cui, Yutong Lu, and Furu Wei. 2022.
\newblock Layoutlmv3: Pre-training for document ai with unified text and image
  masking.
\newblock \emph{arXiv preprint arXiv:2204.08387}.

\bibitem[{Li et~al.(2021)Li, Bi, Yan, Wang, Huang, Huang, and
  Si}]{li2021structurallm}
Chenliang Li, Bin Bi, Ming Yan, Wei Wang, Songfang Huang, Fei Huang, and Luo
  Si. 2021.
\newblock Structurallm: Structural pre-training for form understanding.
\newblock \emph{ACL}.

\bibitem[{Li et~al.(2020)Li, Yin, Li, Zhang, Hu, Zhang, Wang, Hu, Dong, Wei
  et~al.}]{li2020oscar}
Xiujun Li, Xi~Yin, Chunyuan Li, Pengchuan Zhang, Xiaowei Hu, Lei Zhang, Lijuan
  Wang, Houdong Hu, Li~Dong, Furu Wei, et~al. 2020.
\newblock Oscar: Object-semantics aligned pre-training for vision-language
  tasks.
\newblock In \emph{European Conference on Computer Vision}, pages 121--137.
  Springer.

\bibitem[{Liu et~al.(2019)Liu, Ott, Goyal, Du, Joshi, Chen, Levy, Lewis,
  Zettlemoyer, and Stoyanov}]{liu2019roberta}
Yinhan Liu, Myle Ott, Naman Goyal, Jingfei Du, Mandar Joshi, Danqi Chen, Omer
  Levy, Mike Lewis, Luke Zettlemoyer, and Veselin Stoyanov. 2019.
\newblock Roberta: A robustly optimized bert pretraining approach.
\newblock \emph{arXiv preprint arXiv:1907.11692}.

\bibitem[{Luo et~al.(2020)Luo, Ji, Shi, Huang, Duan, Li, Chen, and
  Zhou}]{luo2020univilm}
Huaishao Luo, Lei Ji, Botian Shi, Haoyang Huang, Nan Duan, Tianrui Li, Xilin
  Chen, and Ming Zhou. 2020.
\newblock Univilm: A unified video and language pre-training model for
  multimodal understanding and generation.
\newblock \emph{arXiv preprint arXiv:2002.06353}.

\bibitem[{Nguyen et~al.(2021)Nguyen, Scialom, Staiano, and
  Piwowarski}]{nguyen2021skim}
Laura Nguyen, Thomas Scialom, Jacopo Staiano, and Benjamin Piwowarski. 2021.
\newblock Skim-attention: Learning to focus via document layout.
\newblock \emph{EMNLP Findings}.

\bibitem[{Stanis{\l}awek et~al.(2021)Stanis{\l}awek, Grali{\'n}ski,
  Wr{\'o}blewska, Lipi{\'n}ski, Kaliska, Rosalska, Topolski, and
  Biecek}]{stanislawek2021kleister}
Tomasz Stanis{\l}awek, Filip Grali{\'n}ski, Anna Wr{\'o}blewska, Dawid
  Lipi{\'n}ski, Agnieszka Kaliska, Paulina Rosalska, Bartosz Topolski, and
  Przemys{\l}aw Biecek. 2021.
\newblock Kleister: Key information extraction datasets involving long
  documents with complex layouts.
\newblock \emph{arXiv preprint arXiv:2105.05796}.

\bibitem[{Touvron et~al.(2021)Touvron, Cord, Douze, Massa, Sablayrolles, and
  J{\'e}gou}]{touvron2021training}
Hugo Touvron, Matthieu Cord, Matthijs Douze, Francisco Massa, Alexandre
  Sablayrolles, and Herv{\'e} J{\'e}gou. 2021.
\newblock Training data-efficient image transformers \& distillation through
  attention.
\newblock In \emph{International Conference on Machine Learning}, pages
  10347--10357. PMLR.

\bibitem[{Vaswani et~al.(2017)Vaswani, Shazeer, Parmar, Uszkoreit, Jones,
  Gomez, Kaiser, and Polosukhin}]{vaswani2017attention}
Ashish Vaswani, Noam Shazeer, Niki Parmar, Jakob Uszkoreit, Llion Jones,
  Aidan~N Gomez, Lukasz Kaiser, and Illia Polosukhin. 2017.
\newblock Attention is all you need.
\newblock \emph{Proceedings of the 31st International Conference on Neural
  Information Processing Systems (NeurIPS)}.

\bibitem[{Wang et~al.(2022)Wang, Jin, and Ding}]{wang2022lilt}
Jiapeng Wang, Lianwen Jin, and Kai Ding. 2022.
\newblock Lilt: A simple yet effective language-independent layout transformer
  for structured document understanding.
\newblock \emph{ACL}.

\bibitem[{Xu et~al.(2021{\natexlab{a}})Xu, Xu, Lv, Cui, Wei, Wang, Lu,
  Florencio, Zhang, Che et~al.}]{xu2020layoutlmv2}
Yang Xu, Yiheng Xu, Tengchao Lv, Lei Cui, Furu Wei, Guoxin Wang, Yijuan Lu,
  Dinei Florencio, Cha Zhang, Wanxiang Che, et~al. 2021{\natexlab{a}}.
\newblock Layoutlmv2: Multi-modal pre-training for visually-rich document
  understanding.
\newblock \emph{ACL}.

\bibitem[{Xu et~al.(2020)Xu, Li, Cui, Huang, Wei, and Zhou}]{xu2020layoutlm}
Yiheng Xu, Minghao Li, Lei Cui, Shaohan Huang, Furu Wei, and Ming Zhou. 2020.
\newblock Layoutlm: Pre-training of text and layout for document image
  understanding.
\newblock In \emph{Proceedings of the 26th ACM SIGKDD International Conference
  on Knowledge Discovery \& Data Mining}, pages 1192--1200.

\bibitem[{Xu et~al.(2021{\natexlab{b}})Xu, Lv, Cui, Wang, Lu, Florencio, Zhang,
  and Wei}]{xu2021layoutxlm}
Yiheng Xu, Tengchao Lv, Lei Cui, Guoxin Wang, Yijuan Lu, Dinei Florencio, Cha
  Zhang, and Furu Wei. 2021{\natexlab{b}}.
\newblock Layoutxlm: Multimodal pre-training for multilingual visually-rich
  document understanding.
\newblock \emph{arXiv preprint arXiv:2104.08836}.

\bibitem[{Zaheer et~al.(2020)Zaheer, Guruganesh, Dubey, Ainslie, Alberti,
  Ontanon, Pham, Ravula, Wang, Yang et~al.}]{zaheer2020big}
Manzil Zaheer, Guru Guruganesh, Kumar~Avinava Dubey, Joshua Ainslie, Chris
  Alberti, Santiago Ontanon, Philip Pham, Anirudh Ravula, Qifan Wang, Li~Yang,
  et~al. 2020.
\newblock Big bird: Transformers for longer sequences.
\newblock \emph{Advances in Neural Information Processing Systems (NeurIPS)},
  33.

\end{thebibliography}
